\documentclass[10pt,twocolumn,letterpaper]{article}

\usepackage{cvpr}              

\usepackage{graphicx}
\usepackage{amsmath}
\usepackage{amssymb}
\usepackage{booktabs}

\usepackage[pagebackref,breaklinks,colorlinks]{hyperref}

\usepackage[capitalize]{cleveref}
\crefname{section}{Sec.}{Secs.}
\Crefname{section}{Section}{Sections}
\Crefname{table}{Table}{Tables}
\crefname{table}{Tab.}{Tabs.}

\usepackage{amsmath,amsthm,amsfonts,amssymb,amscd}
\usepackage{graphicx}

\usepackage{multirow}
\usepackage{pifont}
\usepackage{xcolor}         

\newcommand{\cmark}{\ding{51}}%

\newcommand{\boldstart}[1]{\vspace{0.08in}\noindent{\bf #1}}

\def\cvprPaperID{10667} 
\def\confName{CVPR}
\def\confYear{2023}

\begin{document}


\title{StegaPos: Preventing Unwanted Crops and Replacements\\ with Imperceptible Positional Embeddings}


\author{Gokhan Egri and Todd Zickler\\
Harvard University\\
{\tt\small gegri@g.harvard.edu, zickler@seas.harvard.edu}
}


\maketitle

\begin{abstract}

We present a learned, spatially-varying steganography system that allows detecting when and how images 
have been altered by cropping, splicing or inpainting after publication. The system comprises a learned encoder that imperceptibly hides distinct positional signatures in every local image region before publication, and an accompanying learned decoder that extracts the steganographic signatures to determine, for each local image region, its 2D positional coordinates within the originally-published image. Crop and replacement edits become detectable by the inconsistencies they cause in the hidden positional signatures. Using a prototype system for small $(400 \times 400)$ images, we show experimentally that simple CNN encoder and decoder architectures can be trained jointly to achieve detection that is reliable and robust, without introducing perceptible distortion. This approach could 
help individuals and image-sharing platforms certify that an image was published by a trusted source, and also know which parts of such an image, if any, have been substantially altered since publication. 

\end{abstract}

\section{Introduction}

Any image that is shared online is subject to a risk of tampering, because instead of re-sharing a faithful copy of the original, an adversary can share an altered version that has had some pixels replaced with other content (via inpainting or splicing) or that has been cropped to change its meaning (\eg,~\cite{Kennerly.2009, Observers.2018}). The widespread possibility of such tampering creates a fundamental lack of trust: When viewing an image online, an observer cannot know whether it is a faithful rendition of the creator's original intent. 

In the past it has been possible to detect crop, splice and inpainting alterations from the subtle inconsistencies they produced, such as  implausible semantic layouts, inconsistencies in lighting or color tone, and inconsistencies in spatial noise patterns~(\eg,~\cite{zhou2018learning, 8953774, 10.1007/978-3-030-58589-1_19,yerushalmy2011digital,vanhoorick2020dissecting}). But the utility of these approaches has quickly diminished with the rise of modern tampering techniques and AI-based generation technology, which can increasingly replace and create pixels with very few measurable inconsistencies.

We present a new approach for reducing the risk of tampering based on learned steganography. As in conventional deep watermarking and steganography (\eg,~\cite{Baluja2017deep,hayes2017generating,tang2017automatic}), we train an encoder network to embed imperceptible information into an image, and we train an accompanying decoder network to retrieve the information. But instead of embedding one message globally into an image, we design our networks to embed and retrieve \emph{distinct information in every local region}. 
This allows detecting which parts of an image are missing (due to cropping) and which parts have been replaced (due to splicing or generative inpainting).

One way to deploy our networks would be to host them on a secure server or on confidential computing hardware. Trusted content creators---such as journalists, photographers and digital artists---could be authorized to encode their images through an API before publishing them online. Then, upon receiving an image, any individual or image-sharing platform could submit a regulated query to the decoder to determine: (\emph{i}) whether the image can be certified as having originated from the authorized pool of creators; and if so, (\emph{ii}) which parts of the image, if any, have been substantially altered since publication.

We call our system \emph{StegaPos} because the information that it steganographically embeds within each local image region is an encoding of that region's 2D position within the original image. The decoder is trained to recover the hidden positional field by analyzing each local image region, which allows crops and replacements to be detecting from the disturbances they cause in the hidden positional field. Using a large collection of small $(400\times400)$ images, we provide a proof-of-concept demonstration that it is possible to implement this approach using relatively simple CNNs. By training them jointly, we create an encoder-decoder pair that imperceptibly provides useful detection capabilities. In particular, we show these capabilities can be made to robustly survive through common distortions---such as tone-adjustment and resizing---that may be deemed socially ``allowable'' to facilitate convenient sharing between diverse devices without substantially altering the creator's intent. 

Our experiments show quantitatively that our model provides competitive performance across a range of existing benchmarks for splice detection, as well as on a new benchmark for crop detection.

\section{Related Work}
\label{sec:related-work}

\if11

\boldstart{Reactive methods for splice and crop detection} are different from ours because they do not require proactively embedding protective information into an image before it is published. Instead, they react by detecting flaws in a tamperer's work, such as inconsistencies in lighting, shadows, tone mapping, sensor noise signatures, and optical signatures like vignetting. The best reactive methods use deep networks that are trained on large datasets of example splices~\cite{zhou2018learning,8953774,10.1007/978-3-030-58589-1_19} or crops~\cite{vanhoorick2020dissecting}, and compared to our approach, they have the important advantage of being applicable to legacy online images that were not proactively protected. However, any reactive approach is fundamentally in an arms race with counter-forensic methods, which can use increasingly sophisticated processing to eliminate their flaws~\cite{8553401}; and recent advances in AI-based image generation suggest this forensic arms race may soon be lost. 

We show experimentally that our proactive approach improves accuracy on existing splice-detection benchmarks compared to reactive techniques, and we further show that we can detect splices---and crops---when absolutely no inconsistencies are present.

\boldstart{Deep watermarking and deep steganography} use learned encoder and decoder networks to imperceptibly hide a bit-string in an image by modifying its pixel values. This can can be used to broadcast a supplemental piece of information, such as a URL or ownership label~\cite{tancik2020stegastamp, zhu2018hidden, zhang2019steganogan, Wengrowski_2019_CVPR, Wu_2018, hayes2017generating}, or to transmit a covert message that is unnoticeable to adversaries~\cite{zhu2018hidden}. One often wants to build redundancy into the encoding, so that the string can survive through distortions of the host image that may occur during transmission and sharing, such as downsampling, JPEG compression, or re-imaging after projection or printing. An established way to achieve such robustness is to include simulated distortions between the encoder and decoder during training~\cite{tancik2020stegastamp,wengrowski2019light,ahmadi2020redmark}. We find that our networks provide reasonable robustness towards common distortions without augmented training so we leave this for future work.


At a basic level, any watermarking or steganographic system must strike a balance between: (\emph{i}) the amount of hidden information, (\emph{ii}) the perceptible distortion it creates, and (\emph{iii}) its ability to survive distortions~\cite{info11020110}. The existing work has explored learning this balance when a single message is being embedded across all of an image's pixels. In contrast, we explore learning this balance when the message is spatially-varying, meaning that distinct information is being hidden in every local region.

\boldstart{Block-level protection} refers to a class of watermarking techniques that are spatially-varying like ours and that can also protect against replacement alterations. These techniques are cleverly engineered instead of learned. They typically operate by partitioning an image into blocks and modifying some of the bits to embed inter-block signatures that allow detecting when blocks are modified, and furthermore allow recovering contents in the modified blocks (\eg,~\cite{LEE20083497, RePEc:gam:jmathe:v:7:y:2019:i:10:p:955-:d:275854, LIN20052519}). These methods can be very effective (see the supplement for an example) but they are extremely sensitive to distortions like compression and downsampling~\cite{https://doi.org/10.48550/arxiv.1111.6727}, which can limit their utility for practical sharing across devices. Our learning-based approach has a different motivation. It forgoes the ability to reconstruct content that has been modified, and it instead focuses on finding a balance between robustness, protection, and imperceptibility.

\boldstart{Digital signatures} use encryption to attach metadata to an image that allows the recipient to know who created it (attribution) and to know that it has not been altered in transit (exact integrity). Our system serves a different purpose. It is designed to support attribution to an authorised pool of creators, but not to a specific individual within the pool. Also, instead of indicating exact integrity, it provides a softer indication of authenticity to the creator's intent. This has the advantages of being able to survive through distortions like downsampling that  are common to online sharing, and it also provides richer information about where and how an image has been altered since publication.
\fi


\if01

\boldstart{Digital image steganography and watermarking.} Digital image steganography refers to the act of hiding and recovering data in digital images while preserving visual quality~\cite{tancik2020stegastamp, zhu2018hidden, zhang2019steganogan, Wengrowski_2019_CVPR, Wu_2018, 8017430, hayes2017generating, 46526}. Digital image watermarking is a sub-category of digital image steganography which involves encoding a watermark signal (e.g. image, text, binary) in the bits of the original image which can then be decoded for the purposes of authentication, copyright protection, tamper detection, and more \cite{info11020110}.

The main trade-off in digital image watermarking occurs between capacity, the amount of information that can be encoded into the image, robustness, how resilient the encoding is to external image perturbations, and imperceptibility, how invisible the encoding is to the naked eye \cite{info11020110}. Different use cases require a different combination of these attributes, which has given rise to a number of different watermarking schemes such as fragile, semi-fragile, robust, invisible, visible, and more.

\paragraph{Fragile watermarking for authentication and block-level splice detection.} Fragile watermarking is primarily used for tamper detection in images, and is designed such that it is vulnerable to both intentional (e.g. attacks) and unintentional (e.g. gamma compression/expansion, brightness/contrast adjustment, JPEG compression) manipulations. As such fragile watermarking can be thought of as a high capacity, high imperceptibility, low robustness watermarking scheme. The mechanism often consists of a mapping algorithm which maps $m$-by-$n$ pixel blocks in the original image to one another such that information required to recover one block is contained in another block in the same image, a bit encoding scheme which generates a watermark message for each block using the bit-scale information contained in the $m \times n$ pixels in the block and a recovery scheme which allows for the recovery of blocks whose associated recovery blocks are intact and can be used to extract the encoded information about the original block \cite{LEE20083497, RePEc:gam:jmathe:v:7:y:2019:i:10:p:955-:d:275854, LIN20052519}. 

Unlike early watermarking authentication methods which verify the encoded image as a whole, this scheme allows for verifying image authenticity at the block-level, meaning that the model is able to predict whether or not each block of pixels is authentic or tampered with. This makes it possible to detect and localize image tamperings such as splicing, which can be thought of as perturbing only the blocks which intersect with the spliced segment. The primary shortcoming of fragile watermarking is that it is extremely prone to many forms of perturbations, including JPEG compression, blur, color manipulation, noise, and more \cite{https://doi.org/10.48550/arxiv.1111.6727} which makes it an infeasible method for image tampering detection and localization in the real world.



\paragraph{Robust steganography and robust watermarking for authentication.} Robust watermarking refers to the class of watermarking schemes that are designed to withstand many of the possible image perturbations before the encoded image becomes unusable for the purposes of the decoder \cite{TAO2014122}. A trainable extension to robust watermarking is to learn the optimal scheme for encoding and decoding messages in images through data by learning the weights of a neural network encoder and decoder pair. The robustness in this setting is generally satisfied by placing an artificial image perturbation pipeline between the encoder and the decoder, which simulates the perturbations that the decoder needs to be robust against. Common examples of trained against perturbations include camera display transfer function (CDTF) \cite{8954203}, perspective warp \cite{tancik2020stegastamp}, motion defocus blur \cite{tancik2020stegastamp}, color manipulation \cite{tancik2020stegastamp}, noise \cite{tancik2020stegastamp}, JPEG compression \cite{tancik2020stegastamp}. In order to be robust against real-life conditions, these architectures are designed with redundancy in encoding such that the entirety of the message can be recovered with only a portion of the image intact. As such, robust watermarking can be considered a high robustness, high imperceptibility, low capacity watermarking scheme. In addition, the redundant nature of the encoding also ensures that the encoding is non-localized, meaning that different image patches in the encoded image can not be differentiated through their encoding alone, which is required for crop and splice detection. 

\boldstart{Splice detection.} One can often detect splices passively, without pro-actively hiding signatures like we do, by exploiting anomalies in the semantic layout of an image and by exploiting inconsistencies in lighting, shadows, tone mapping, camera noise signatures, and so on. The most effective methods in this class train deep neural networks on large datasets of example splices~\cite{zhou2018learning,8953774,10.1007/978-3-030-58589-1_19} and produce either splice bounding boxes or pixel-level splice masks. 
Our model also produces pixel-level splice masks, but because it uses pro-active encoding, it improves performance and is less likely to be fooled in the future as tamperers and their counter-forensic tools continue advancing.

\boldstart{Crop detection.} Passively detecting crops is much harder because there are fewer anomalies and inconsistencies to exploit. Semantic anomalies are weak because it is hard to distinguish between (i) a region that was cropped from a published image and (ii) a similar authentic image that was framed and published that way by a photographer. As a result, techniques in this category produce relatively coarse crop localizations, and they rely on the presence of radial optical effects like chromatic aberration, radial distortion and vignetting~\cite{yerushalmy2011digital,vanhoorick2020dissecting}. In contrast, our steganographic approach is more precise and succeeds whether these effects are present or not. For example, it can be used to protect the published assets of digital content creators in addition to photographers' images. Crop detection is also an ill-posed problem for existing digital image watermarking and trainable robust watermarking methods, as distinguishing between two sub-patches of the same original image requires the encoded information to be localized and non-redundant at the patch level, which is not satisfiable by either methods. Our proposed method is able to carry out both crop detection and localization, by locally-encoding position and scale information on each patch which can then be decoded to distinguish between two different patches of the same image.



\fi

\section{Steganographic Positional Embedding}

\begin{figure*}[t]
  \centering
  \includegraphics[width=\textwidth]{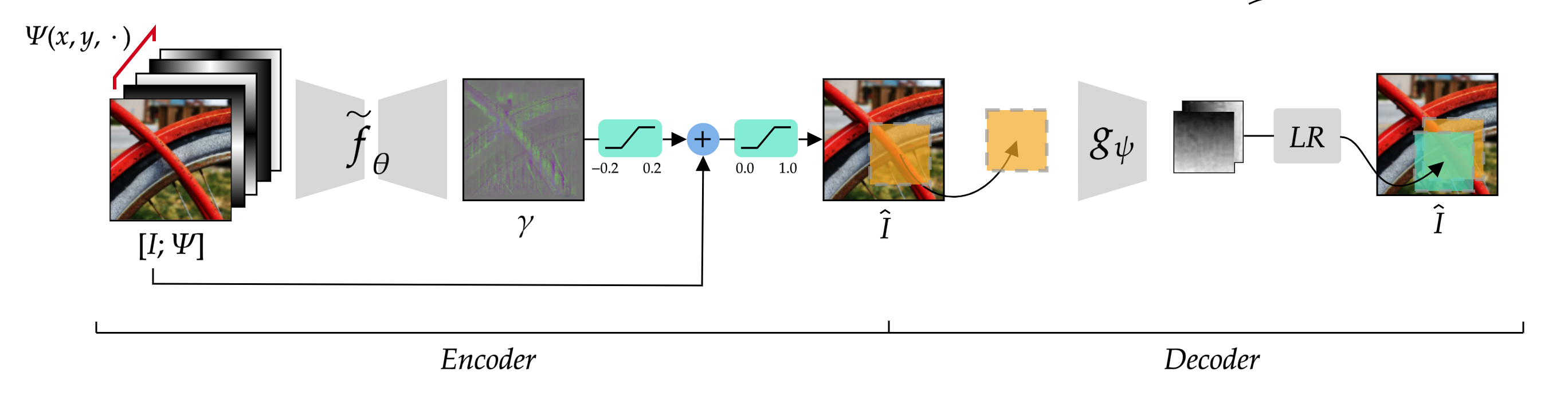}\vspace*{-0.2in}
  \caption{System overview. The encoder concatenates input image $I(x,y)$ with frequency-based positional codes $\Psi(x,y,\cdot)$ and maps these to a residual signal $\gamma(x,y)$ that is added to $I$ (with clamping) to produce stegapos image $\hat{I}$. The decoder receives a (possibly cropped and downsampled) image $\hat{I}(i,j)$ and maps this to an array of 2D positions $p(i, j)$ representing the pixel coordinates at which the content at $(i,j)$ was located within the originally-encoded image. The decoded positional field $p(i,j)$ is sufficient to determine whether the input $\hat{I}$ is stegapos-encoded, and if so, which regions have been altered by replacements. Additionally, by passing the field to a subsequent linear regression operator (LR, see Eq.~\ref{eqn:linear_reg_task}), the system infers the top-left offset of a crop and/or a downsampling factor.}
  \label{fig:encoder_architecture}
\end{figure*}

Our system is shown in Figure~\ref{fig:encoder_architecture}. A color image $I(x,y,c)\in [0, 1]$ of size $H\times W\times 3$ is input to a CNN $\tilde{f}_\theta$ which generates a signal $\gamma(x,y,c)$ of the same size. The signal is added (with clamping) to create an encoded image $\hat{I}$ that is perceptually close to $I$. We call $\hat{I}$ a steganographic-positional image, or \emph{stegapos} image for short. It is meant to contain, within each local image patch, sufficient information about the spatial position $p=(x,y)\in \{1\ldots W\}\times\{1\ldots H\}$ of the patch's center.

Later, when a stegapos image $\hat{I}(i,j,c)\in [0, 1]$ arrives at the decoder, possibly after being cropped or downsampled to a different size, a CNN $g_\psi$ maps it to an array of inferred position values $p(i,j)\in [1,W]\times [1,H]$, and then a linear regression operator (LR) estimates from $p(i,j)$ a decimation scale-factor and the top-left offset of a crop. 

The overarching idea is very simple. In the absence of any splicing, cropping or downsampling, the decoder's network $g_\psi$ should recover from $\hat{I}$ a positional field that is close to the one that was embedded: $p(i,j)\approx(i,j)\ \forall (i,j)\in \{1\ldots W\}\times\{1\ldots H\}$. If there is cropping or downsampling applied to $\hat{I}$, the recovered positional field will also be cropped or downsampled, and this will be detected by the linear regression operator. Moreover, if there is a replacement applied to $\hat{I}$, it causes an anomaly in the positional field (\eg, Figure~\ref{fig:raw_decoder_outputs}) that can also be detected.

In what follows, our prototype system assumes that all of the encoded images are of a single, pre-determined size ($H\times W$), and that images sent to the decoder are of equal size or smaller. Extensions to larger and variable image sizes are left for future work.

\subsection{Encoding} \label{encoder}

We use a standard U-Net~\cite{long2015fully, ronneberger2015unet} for $\tilde{f}_\theta$ with input that is a channel-wise concatenation of the image $I$ and an $H\times W \times D$ array of frequency-based positional codes,
\begin{align*}
    \Psi(x,y,\cdot) &= \left\{ \begin{bmatrix} \cos(\omega_k x) \\ \sin(\omega_k x) \\ \cos(\omega_k y) \\ \sin(\omega_k y) \end{bmatrix} \mid \omega_k=\omega_o^{4k/D}, k=1\ldots \tfrac{D}{4} \right\},
\end{align*}
with frequencies $\omega_k$ forming a geometric progression from sufficiently small base frequency $\omega_o$. In our experiments we use dimension $D=8$ and follow~\cite{vaswani2017attention} by using base frequency $\omega_o=10^{-4}$.

The CNN output is then clamped and added to the input image according to
\begin{align}
    \gamma &= \tilde{f}_{\theta}(\begin{bmatrix} I; \Psi \end{bmatrix}) \\
    \hat{I} &= \text{clamp}_{[0, 1]}(I + \text{clamp}_{[-0.2, 0.2]}(\gamma)).
\end{align}
We use notation $\hat{I}=f_{\theta}(I)$ for the complete mapping from input to stegapos output.

The frequency-based codes $\Psi$ play an important role and are a convenient input for several reasons. In addition to having values in $[-1,1]$ and providing a unique $D$-dimensional signature at each position $(x,y)$, they provide the CNN with a range of input spatial frequencies that it can mix to create local residuals that ``match'' (in a perceptual sense) the various local textures and colors of an input image. The codes also have a linear and shift-invariant relational property, meaning that codes $\Psi(x,y,\cdot)$ and $\Psi(x+\Delta x,y+\Delta y,\cdot)$ are related by a linear transformation (see expression in the supplement) that depends only on $(\Delta x, \Delta y)$. This can help to ``work around'' smooth, uniformly-colored parts of an input image, which tend to be perceptually sensitive to rapidly-varying residuals, by instead placing information about relative position at nearby points that are less perceptually sensitive.

\subsection{Decoding} \label{decoder}
The decoder receives a color image $\hat{I}(i,j,c)\in[0,1]$ of size $N\times M\times 3,$ with $N\le W, M\le H$. We use a standard, undecimated (\ie, stride-$1$) CNN for $g_\psi$ to map from the image to its embedded positional field $p(i,j)$. We discard positional estimates near the borders of the image that are affected by windowing, so the CNN's output is of size $(N-R+1)\times (M-R+1)$, where $R$ is the receptive field size,  and its output values $p$ are nominally in the range $[\lfloor \frac{R}{2} \rfloor, W - \lceil \frac{R}{2} \rceil] \times [\lfloor \frac{R}{2} \rfloor, H - \lceil \frac{R}{2} \rceil]$.


The receptive field $R$ is determined by the CNNs filters and layers and is a critical design parameter. Larger receptive fields allow aggregating positional information about each $(i,j)$ from a larger neighborhood around it, which increases the accuracy of the positional estimates and, perhaps more importantly, increases the local information capacity that is available to the encoder for embedding positional information with less perceptual distortion. On the other hand, smaller receptive fields increase the spatial precision of the positional estimates, allowing the detection of positional aberrations (due to crops or splices) that are smaller in size. In our experiments, we use $R=50$, which is $12.5\%$ of the image width.



The network $g_\psi$ is trained, via augmentation as described below, to be approximately decimation-equivariant. That is, if $p=g_\psi(I)$ and $q=g_\psi(I_{\downarrow_{1/s}})$ for some decimation factor $s\in\left[s_\text{min}, 1\right]$ in a pre-determined range, then the two outputs are related by $p_{\downarrow_{1/s}} \approx q$. This ensures that hidden positional signatures can be recovered in spite of downsampling that occurs between encoding and decoding, and, as a by-product, it also allows detecting when a stegapos image has been downsampled and by how much.

The linear regression operator accepts the estimated positional field $p(i,j)=(p_x(i,j),p_y(i,j))$ and analytically computes least squares estimates of the decimation scale factor and the top-left coordinates of a crop. An image cropped with top-left offset $(\mu_x,\mu_y) \in [1, W]\times [1, H]$ and scaled by $s\in\left[s_\text{min}, 1\right]$ will have ideal CNN outputs
\begin{align}
\mathcal{P}^{([\mu_x, \mu_y], s)}_{x}(i, j) \triangleq \mu_x + (\lfloor{\tfrac{R}{2}}\rfloor + i)\tfrac{1}{s} \\
    \mathcal{P}^{([\mu_x, \mu_y], s)}_{y}(i, j) \triangleq \mu_y + (\lfloor{\tfrac{R}{2}}\rfloor + j)\tfrac{1}{s},
    \label{eqn:positional-fields}
\end{align}
so the linear regression operator estimates the offset ($\mu_x$, $\mu_y$) and scale $s$ via
\begin{align}
    \hat{\mu}_x, \hat{s}_x &= \sum_{ij} \text{arg}\min_{\mu_x, s} ||p_x(i,j) - \mathcal{P}^{([\mu_x, \mu_y], s)}_{x}(i,j)||_2^2 \\
    \hat{\mu}_y, \hat{s}_y &= \sum_{ij} \text{arg}\min_{\mu_y, s} ||p_y(i,j) - \mathcal{P}^{([\mu_x, \mu_y], s)}_{y}(i,j)||_2^2 \\
    \hat{s} &= \frac{\hat{s}_x + \hat{s}_y}{2}.
    \label{eqn:linear_reg_task}
\end{align}
\noindent Note that $\mathcal{P}^{([0, 0], 1)}$ is the expected CNN output for an image that has not been cropped or scaled. Also note that it would be possible to separately estimate $\hat{s}_x$ and $\hat{s}_y$ in order to detect scalings of aspect ratio.

\subsection{Training}

We train the system end-to-end with a loss that simultaneously promotes positional accuracy and visual fidelity:
\begin{align}
\label{eqn:loss_function}
\begin{split}
    \mathcal{L}(\theta,\psi,\phi) &= ||p - \mathcal{P}^{([0, 0], s)}_{x, y}||_2^{2} 
     + \lambda_{\gamma} ||\gamma||_1 \\
    & + \lambda_{I} ||\hat{I}_{YUV} - I_{YUV}||_1 + \lambda_{s} \mathcal{L}_{s}(I, \hat{I})  \\
    & - \lambda_{c}\log d_\phi(f_{\theta}(I))
\end{split}
\end{align}
\noindent Here, $\gamma=\tilde{f}_\theta(I)$ is the residual image, $\hat{I}=f_\theta(I)$ is the encoded image, $p=g_\psi(f_\theta(I))$ is the decoded positional field, and $I_{YUV}$ is an image converted to YUV color space. $\mathcal{L}_{s}(I, \hat{I})$ is a perceptual loss~\cite{DBLP:journals/corr/abs-1801-03924}, and the final term, with weight $\lambda_c$, is a critic loss that employs an auxiliary CNN discriminator $d_\phi: \{I\}\mapsto [0, 1]$ with weights $\phi$ that adapt to differentiate between input and encoded images~\cite{larsen2016autoencoding}. We alternate between updating weights $\phi$ to reduce $\mathcal{L}_{D} = -\log d_\phi(I) - \log (1 - d_\phi(\hat{I}))$ and updating weights $\theta,\psi$ to reduce $\mathcal{L}(\theta,\psi)$.

To achieve approximate decimation-equivariance in the decoder, we find it sufficient to use a simple augmentation approach during training: We randomly scale each encoded image by $s \in [s_\text{min}, 1]$ before feeding it to the decoder and then evaluate the positional loss between the outputs $p$ and the expected decimated output $\mathcal{P}^{([0, 0], s)}_{x, y}$. 

We use a two-phase training regime similar to \cite{tancik2020stegastamp}, first training for positional accuracy by setting $\lambda_{I}=\lambda_{c}=\lambda_{s}=\lambda_{\gamma}=0$, and then training for both position and visual quality using all loss terms until convergence. We find the first phase converges quickly with high positional accuracy but low visual quality, and that visual quality restores when the regularization coefficients $\{\lambda_{I}, \lambda_{c}, \lambda_{\gamma},\lambda_{s}\}$ are gradually increased during the second phase.

\section{Experiments}

We train using 100,000 images of size $400 \times 400$ from the MIRFLICKR 1M dataset. 
We use dimension $D = 8$ with base frequency $\omega_o=10^{-4}$ for the input positional codes. The decoder receptive field size is $R = 50$, and we train over decimation factors $s\in[0.2, 1]$.  Architectural details for the three CNNs are in the supplement.

During training, we use a batch size of $8$ images and the Adam optimizer with learning rate $10^{-7}$ and $(\beta_1, \beta_2) = (0.9, 0.999)$. Overall, we find that training is quite susceptible to poor local minima, and we resort to manual manipulation of coefficients $\{\lambda_{I}, \lambda_{c}, \lambda_{\gamma},\lambda_{s}\}$ during training. 

Figure~\ref{fig:raw_decoder_outputs} visualizes a residual signal created by the encoder along with the positional field recovered by the decoder. Figures~\ref{fig:qualitative_crop_results} and \ref{fig:qualitative_splice_results} show some representative results for applications to crop and replacement detection. Overall, the following experiments show the system is capable of preserving high visual quality while also achieving high positional accuracy. It provides close equivariance with decimation, is robust to several allowable distortions, and enables the detection of splice/replacement masks with useful spatial precision. At the end of this section, in Part~\ref{sec:statistics}, we provide insight into what the system has learned by examining some statistics of the residual signals created by the encoder.

\subsection{Stegapos or not?}
\label{net:stegaposdetect}
We begin with the authentication task of classifying an image as being stegapos-encoded or not. This is useful to an observer who wants to verify the source of the image.

\begin{figure*}[t]
  \centering
  \includegraphics[width=\textwidth]{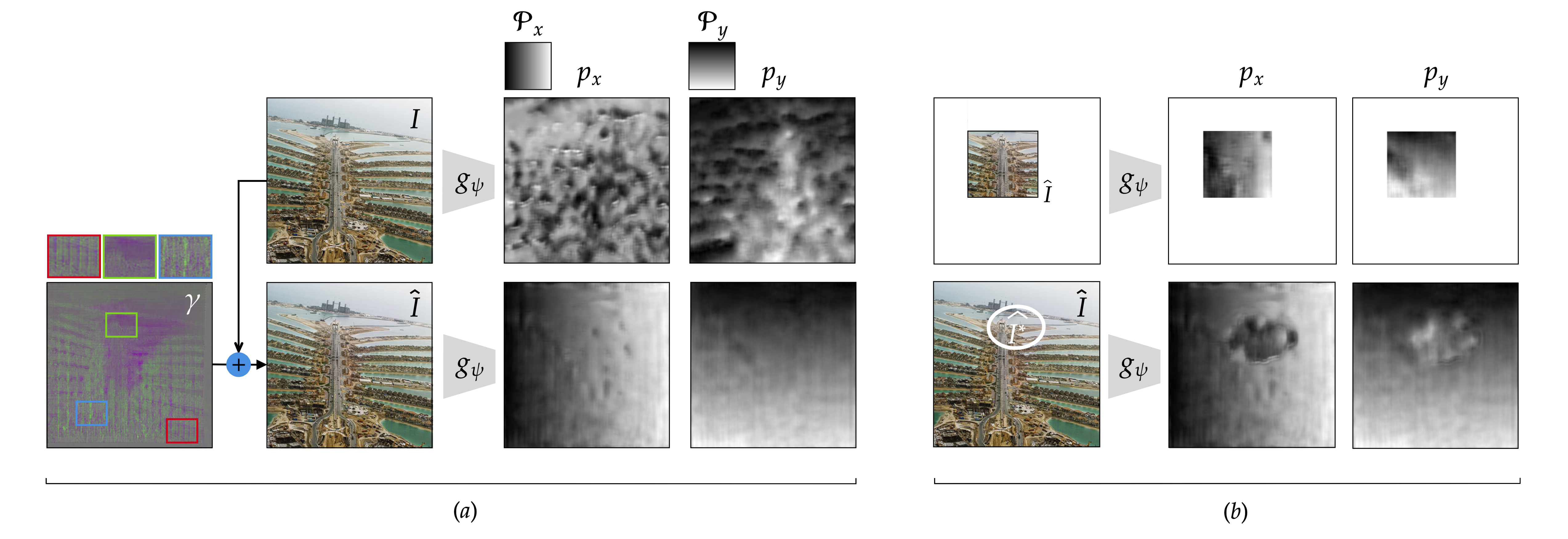}
  \caption{Visualizations of decoded positional fields $p(i,j)=\left(p_x(i,j),p_y(i,j)\right)$. (a) An unencoded image $I$ maps to a structureless field while an encoded one $\hat{I}$ maps close to smooth positional field $\left({\cal P}_x,{\cal P}_y\right)$ that was embedded. (b) Examples of positional fields decoded after crops or replacements. Anomalies in the field caused by a replacement are detectable even when the replacement is visually perfect. See text for details.}
  \label{fig:raw_decoder_outputs}
\end{figure*}

The visualizations in Figure~\ref{fig:raw_decoder_outputs}(a) suggest that stegapos images $\hat{I}$ can be easily distinguished by the positional estimates $p=g_\psi(\hat{I})$ they induce. To verify this, we train a single layer classifier with sigmoid activation that accepts $p(i,j)$ and estimates the probability that $\hat{I}$ is stegapos-encoded.
We train and validate the classifier on a 25,000/10,000-split of the original MIRFLICKR 1M dataset, applying stegapos-encoding to half of the images and leaving the rest unencoded. We freeze the encoder and decoder parameters and optimize the classifier weights using the $\ell_2$-loss between the estimated and ground-truth binary (encoded/unencoded) labels. The classifier converges quickly and provides $100\%$ training accuracy and $99.4\%$ validation accuracy, from which we conclude that verification of stegapos-encoding is reliably achievable.


\subsection{Detecting crops and downsampling}  
\begin{figure}[t]
  \centering
  \includegraphics[width=8cm]{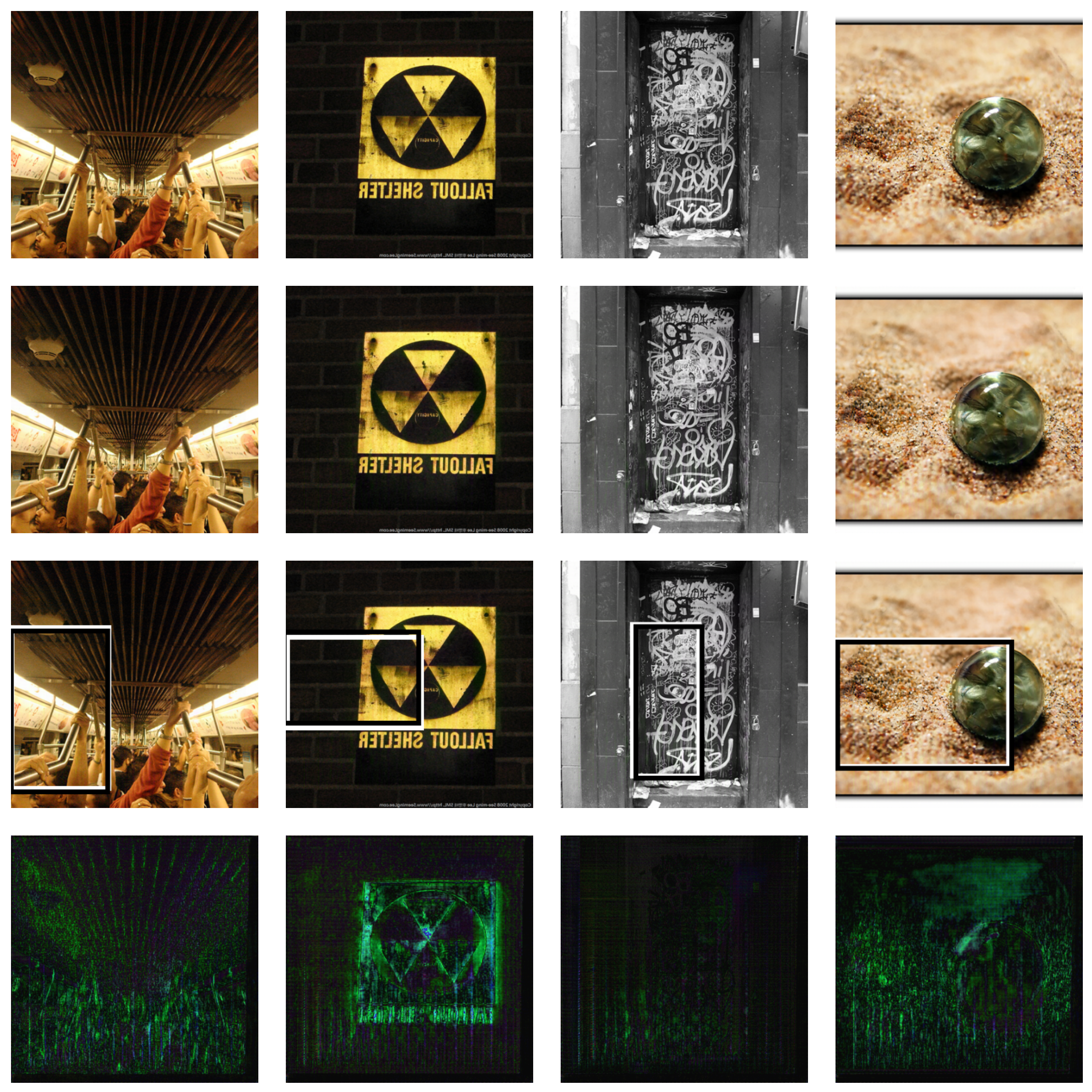}
  \caption{Sample crop detection results ($s=1.0$). (top to bottom) Original image $I$, stegapos image $\hat{I}$, ground truth crop (black) and estimated crop (white), residual image $|\hat{I} - I|$ with values scaled $\times 10$ for visibility.
  }
  \label{fig:qualitative_crop_results}
\end{figure}

Next we consider the task of estimating the scale and crop-offset of a stegapos image by using the offset and scale estimates $(\hat{\mu}_x,\hat{\mu}_y, \hat{s})$ from Equation \ref{eqn:linear_reg_task}. Figure~\ref{fig:qualitative_crop_results} shows typical results, where the model is able to accurately estimate the cropped regions without sacrificing image quality. 

\begin{figure}[t]
  \centering
  \includegraphics[width=8cm]{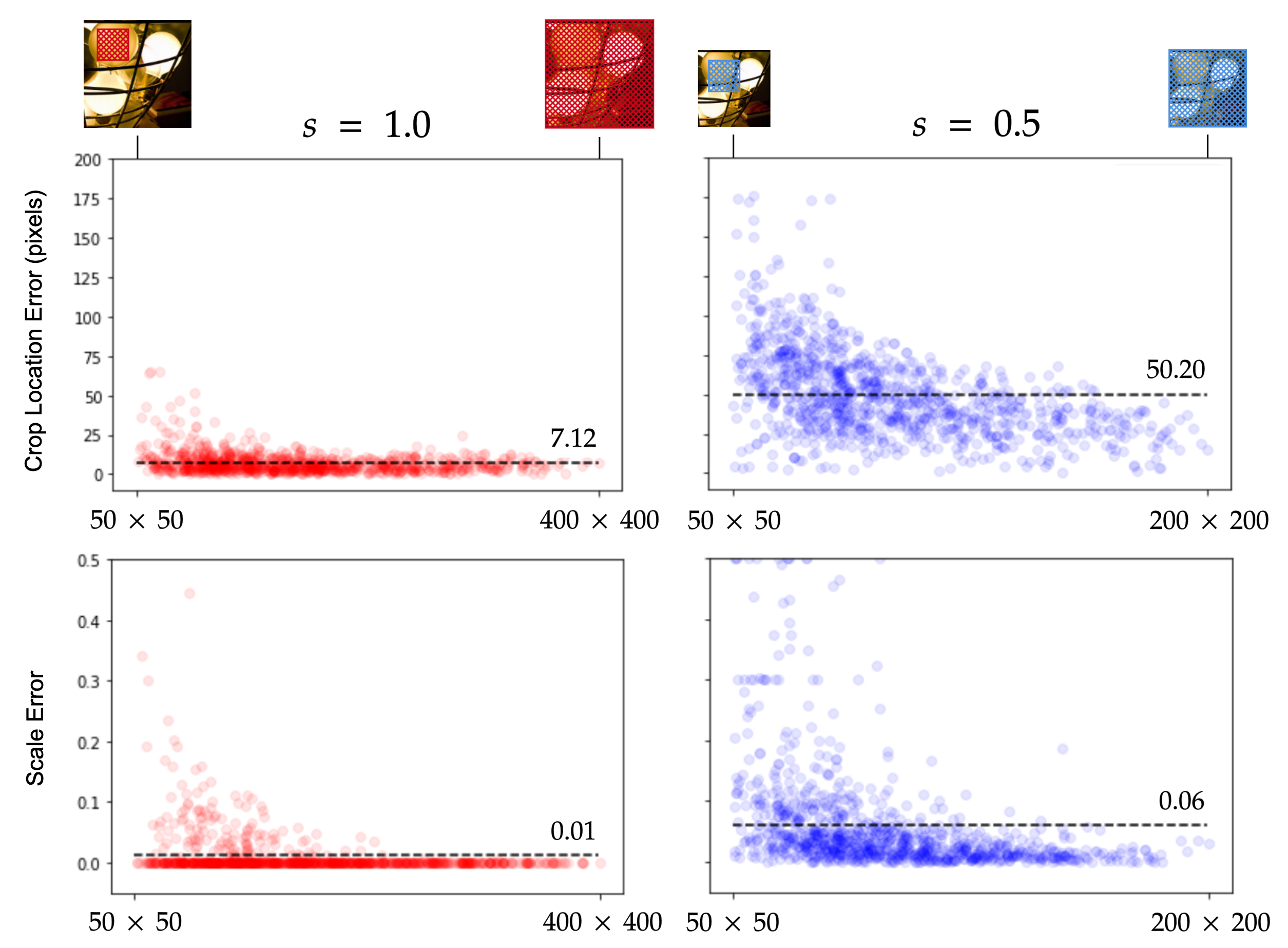}
  \caption{Crop detection errors versus crop size, from $50 \times 50$ ($1/16$ crop) to $400 \times 400$ (no crop), visualized using 1000 random samples from our 25,000-image dataset. Columns show errors for two scales, $s \in \{1.0, 0.5\}$. Top: Error in estimated top-left corner $(\mu_x,\mu_y)$, measured in pixels. Bottom: Error in estimated scale $s$. For more complete results, please check the supplement.}
  \label{fig:quantitative_crop_results}
\end{figure}

To quantify accuracy, we build a new dataset we call \texttt{SmartCrop25K}, comprising 25,000 images of size $400 \times 400$ from the MIRFLICKR 25K dataset that are cropped at various sizes from full size (no crop) down to $1/16$-size ($50 \times 50$ crop). The cropping was done automatically by a saliency-aware system that preserves salient content. Figure~\ref{fig:quantitative_crop_results} visualizes our model's quantitative results for offset and scale on this dataset. 

Crop localization is harder when an image is downsampled ($s<1$) and when the crop retains a smaller fraction of the image. We find that when the input images are not downsampled (first column, $s = 1$) the model's crop offset error is less than 25 pixels for all but the smallest crop sizes. When the images are downsampled (second column) the error degrades in a graceful manner. See the supplement for results on additional values of $s$. 

\subsection{Detecting replacements}
Finally, we consider the task of detecting replacements, where a section of a stegapos image has been replaced by pixels from another image or generated content. We quantify performance using examples of two-factor splices, where a composite image $\hat{I}$ has been created by blending content from two different sources according to $\hat{I}(i,j,c)={\cal M}(i,j)\hat{I}_1(i,j,c)+(1-{\cal M}(i,j))\hat{I}_2(i,j,c)$ with binary mask ${\cal M}\in\{0,1\}$. Our task is to infer the mask ${\cal M}$ from image $\hat{I}$, without prior knowledge of $\hat{I}_1,\hat{I}_2$. There are two cases to consider: (\texttt{ee}) both sources are stegpos-encoded; and (\texttt{eu}) one source is stegapos-encoded and the other source is unencoded.

Figure~\ref{fig:raw_decoder_outputs}(b) visualizes an artificially-challenging example of the  (\texttt{ee}) case, where a composite image is created by copies of itself
that are encoded with different relative positions, that is $\hat{I}={(1-\cal M)}f_\theta(\hat{I})+{\cal M}f^\Delta_\theta(\hat{I})$ with $f^\Delta$ an encoding from use of phase-shifted input positional codes $\Psi^\Delta$. There is absolutely no perceptual evidence for the existence of the replacement, but the anomaly in the recovered positional signature still provides a clear signal for detecting its presence. 

The most direct way to estimate mask ${\cal M}$ from input image $\hat{I}$ is to apply a threshold to the per-pixel deviations following the decoder's linear regression operator, that is 
\begin{align} \label{eqn:splice_threshold_equation}
    \hat{\mathcal{M}}(i, j) = |p(i, j) - \mathcal{P}^{([\hat{\mu}_x, \hat{\mu}_y], \hat{s})}_{x, y}(i, j)| > \alpha,
\end{align}
with some threshold value $\alpha$ and with $\hat{\mu}_x, \hat{\mu}_y, \hat{s}$ 
provided by the linear regression operator. We call this approach ``Ours+L'' (for linear) in Table~\ref{table:splice_f1_table}, and we test situations where (i) the threshold $\alpha$ is fixed across an entire collection of images; and (ii) it is optimally chosen (by an oracle) for each image.

For comparison, we also test a more sophisticated method for splice detection that replaces $g_\psi$ with a new U-Net $h_\phi$, which maps an (undecimated and uncropped) image $\hat{I}(i,j,c)$ directly to a binary mask $\hat{\cal M}(i,j)$. We call this approach ``Ours+N'' (for network) in Table~\ref{table:splice_f1_table}. We train $h_\phi$ using a frozen encoder and a synthetic training set comprising labeled pairs $(\hat{I}^{(i)},{\cal M}^{(i)})$ that are created by combining a 25,000-image subset of the MIRFLICKR 1M dataset with a generated set of simple masks (circles, squares, etc.) using the ($\texttt{ee}$) scheme. The weights $\phi$ are optimized using loss $||\mathcal{M} - h_\phi(\hat{I}))||^2_2$, and they converge very quickly.

\begin{figure}[h]
  \centering
  \includegraphics[width=8cm]{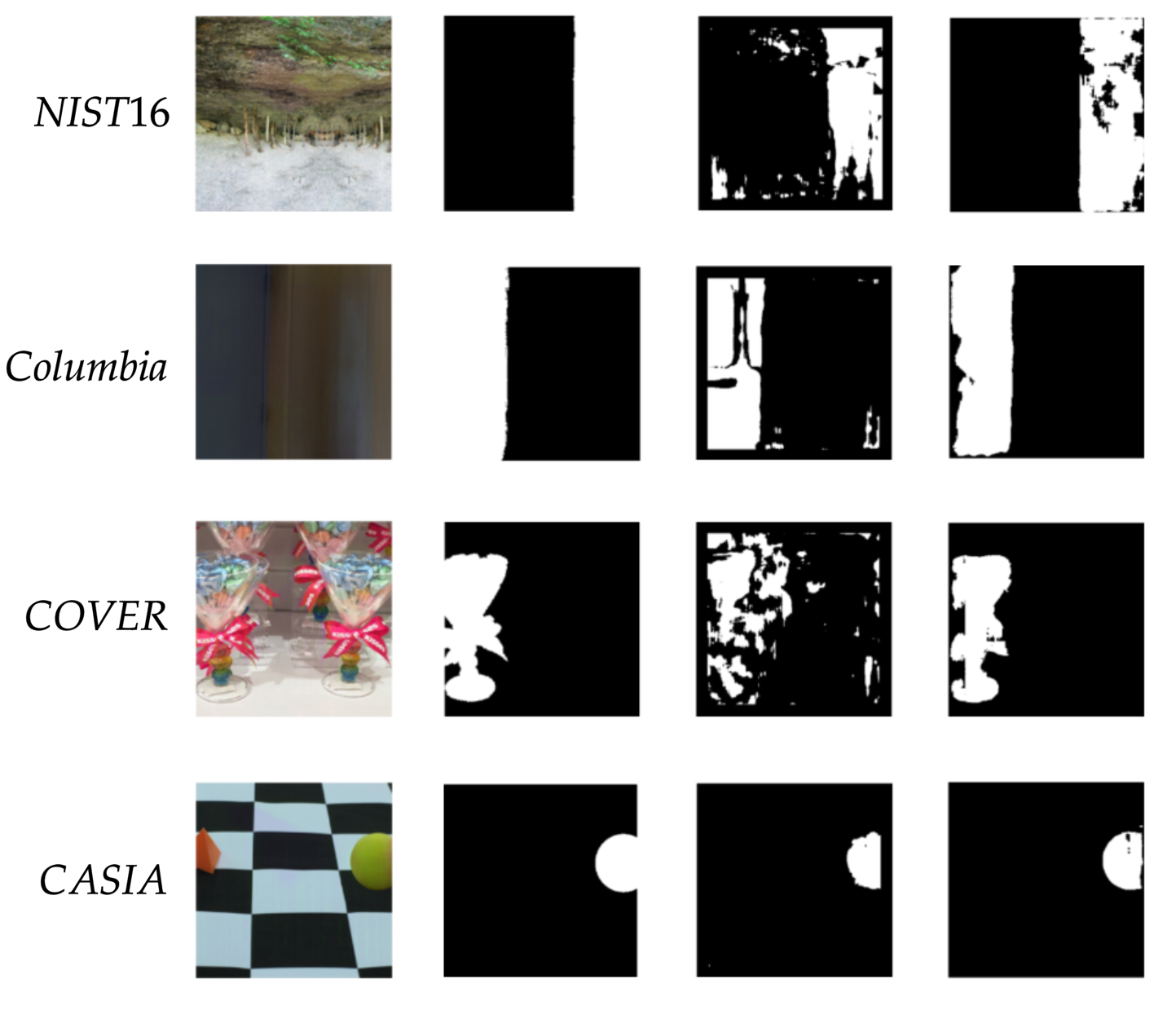}
  \caption{Some of our splice-detection results using examples (in the ($\texttt{ee}$) scheme) taken from four existing benchmarks. From left to right: input image, ground truth mask, estimated mask via per-pixel threshold (Ours+L), estimated mask via post-trained splice network (Ours+N).}
  \label{fig:qualitative_splice_results}
\end{figure}

We evaluate using four existing benchmarks: NIST-16~\cite{nist-16-paper}, Columbia~\cite{hsu06crfcheck}, COVERAGE~\cite{7532339}, and CASIA~\cite{casia-paper-1}. Each benchmark includes labeled pairs $(I,{\cal M})$ that we downsample to $400\times 400$. Figure~\ref{fig:qualitative_splice_results} shows examples and Table~\ref{table:splice_f1_table} provides numbers, both with comparisons to previous methods that do not use proactive embedding (\ie, that are ``reactive'' according to the taxonomy of Section~\ref{sec:related-work}). The table uses superscript F to indicate methods that are fine-tuned separately for each dataset; and superscript O to indicate methods that use an optimal, oracle-provided threshold for each image (analogous to $\alpha$ in Eq.~\ref{eqn:splice_threshold_equation}), which is a common practice for these benchmarks~\cite{zhou2018learning, Salloum_2018, 10.1007/s11042-016-3795-2}.

As expected, our proactive-encoding approach can outperform purely reactive ones. Applying a simple per-pixel threshold after the regression operator (Ours+L) already does better in many cases, and post-training a dedicated splice-detection U-Net $h_\phi$ (Ours+N), even with very simple synthetic data, does even better by exploiting the spatial coherence of mask-shapes. We find that it substantially outperforms all existing splice detectors, without fine-tuning and without oracle thresholding. We also find, for all variations of our model, that detection accuracy is similar across the ($\texttt{ee}$) and ($\texttt{eu}$) situations.

\begin{small}
\begin{table}[t]
\setlength{\tabcolsep}{5pt} 
\small
  \caption{Splice mask $F_1$-scores on four benchmarks. Superscript F indicates a method is fine-tuned separately for each benchmark dataset, and superscript O indicates a method uses an optimal, oracle-provided threshold for each  image within a dataset. Missing entries are not available in prior literature.}
  \centering 
  \begin{tabular}{lcccc}
    \toprule
        Method    & NIST16 & Columbia & COVER. & CASIA \\
    \midrule
     \textsuperscript{\phantom{F}O}ELA \cite{ela-paper}    & 0.236 & 0.470 & 0.222 & 0.214 \\
     \textsuperscript{\phantom{F}O}NOI1 \cite{10.1016/j.imavis.2009.02.001}  & 0.285 & 0.574 & 0.269 & 0.263 \\
     \textsuperscript{\phantom{F}O}CFA1 \cite{6210378}  & 0.174 & 0.467 & 0.190 & 0.207 \\
     \textsuperscript{\phantom{F}O}MFCN \cite{Salloum_2018}  & 0.571 & 0.612 & -     & 0.541  \\
    \textsuperscript{FO}RGB-N \cite{zhou2018learning} & 0.722 & 0.697 & 0.437 & 0.408 \\
    \textsuperscript{F\phantom{O}}SPAN \cite{10.1007/978-3-030-58589-1_19} & 0.582 & - & 0.558 & 0.382 \\
    \midrule
     \textsuperscript{\phantom{FO}}Ours + L (\texttt{eu}) & 0.535 & 0.500 & 0.650 & 0.390 \\
     \textsuperscript{\phantom{F}O}Ours + L (\texttt{eu}) & 0.658 & 0.569 & 0.706 & 0.585 \\
     \textsuperscript{\phantom{FO}}Ours + N (\texttt{eu}) & \textbf{0.835} & \textbf{0.800} & \textbf{0.890} & \textbf{0.704} \\
    \midrule
     \textsuperscript{\phantom{FO}}Ours + L (\texttt{ee}) & 0.526 & 0.535 & 0.639 & 0.388 \\
     \textsuperscript{\phantom{F}O}Ours + L (\texttt{ee}) & 0.659 & 0.606 & 0.719 & 0.595 \\
     \textsuperscript{\phantom{FO}}Ours + N (\texttt{ee}) & \textbf{0.843} & \textbf{0.809} & \textbf{0.881} & \textbf{0.733} \\
    \bottomrule
  \end{tabular}
  \label{table:splice_f1_table}
\end{table}
\end{small}

\if01
\begin{small}
\begin{table}[t]
\small
  \caption{Splice mask $F_1$-scores on four standard datasets. ``Fine-tuned'' indicates models that are exposed and adjusted to each benchmark during training, and ``oracle threshold'' indicates models that use the optimal threshold $\alpha$ for each image. For our model, ``+ Linear'' refers to thresholded linear regression residuals and ``+ Network'' refers to the dedicated splice detection U-Net. Higher scores are better. Missing entries are those not yet reported in existing literature.}
  \centering 
  \scalebox{0.9}{
  \begin{tabular}{>{\centering\arraybackslash}p{12mm}>{\centering\arraybackslash}p{12mm}lcccc}
    \toprule
         FT & OT & Method    & NIST16 & Columbia & COVER & CASIA \\
    \midrule
     & \cmark & ELA \cite{ela-paper}    & 0.236 & 0.470 & 0.222 & 0.214 \\
     & \cmark & NOI1 \cite{10.1016/j.imavis.2009.02.001}  & 0.285 & 0.574 & 0.269 & 0.263 \\
     & \cmark & CFA1 \cite{6210378}  & 0.174 & 0.467 & 0.190 & 0.207 \\
     & \cmark & MFCN \cite{Salloum_2018}  & 0.571 & 0.612 & -     & 0.541  \\
    \cmark & \cmark & RGB-N \cite{zhou2018learning} & 0.722 & 0.697 & 0.437 & 0.408 \\
    \cmark &  & SPAN \cite{10.1007/978-3-030-58589-1_19} & 0.582 & - & 0.558 & 0.382 \\
    \midrule
     &  & Ours + Linear (\texttt{enc-unenc}) & 0.535 & 0.500 & 0.650 & 0.390 \\
     & \cmark & Ours + Linear (\texttt{enc-unenc}) & 0.658 & 0.569 & 0.706 & 0.585 \\
     &  & Ours + Network (\texttt{enc-unenc}) & \textbf{0.835} & \textbf{0.800} & \textbf{0.890} & \textbf{0.704} \\
    \midrule
     &  & Ours + Linear (\texttt{enc-enc}) & 0.526 & 0.535 & 0.639 & 0.388 \\
     & \cmark & Ours + Linear (\texttt{enc-enc}) & 0.659 & 0.606 & 0.719 & 0.595 \\
     &  & Ours + Network (\texttt{enc-enc}) & \textbf{0.843} & \textbf{0.809} & \textbf{0.881} & \textbf{0.733} \\
    \bottomrule
  \end{tabular}
  }
  \label{table:splice_f1_table}
\end{table}
\end{small}
\fi

\subsection{Robustness to Allowable Distortions}
Many sharing practices involve transferring images between devices that have varying communication bandwidths and display resolutions, so it is common---at least when archival quality is not concerned---for images to experience distortions such as tone-adjustment and downsampling. Society often accepts these distortions as ``allowable'' because they provide convenience without substantially altering the intent of the image creator. It is desirable for a system like ours to be insensitive to as many of these distortions as possible, because it enables a greater number of images to be properly certified as being trustworthy.

\begin{figure}[h]
  \centering
  \includegraphics[width=8cm]{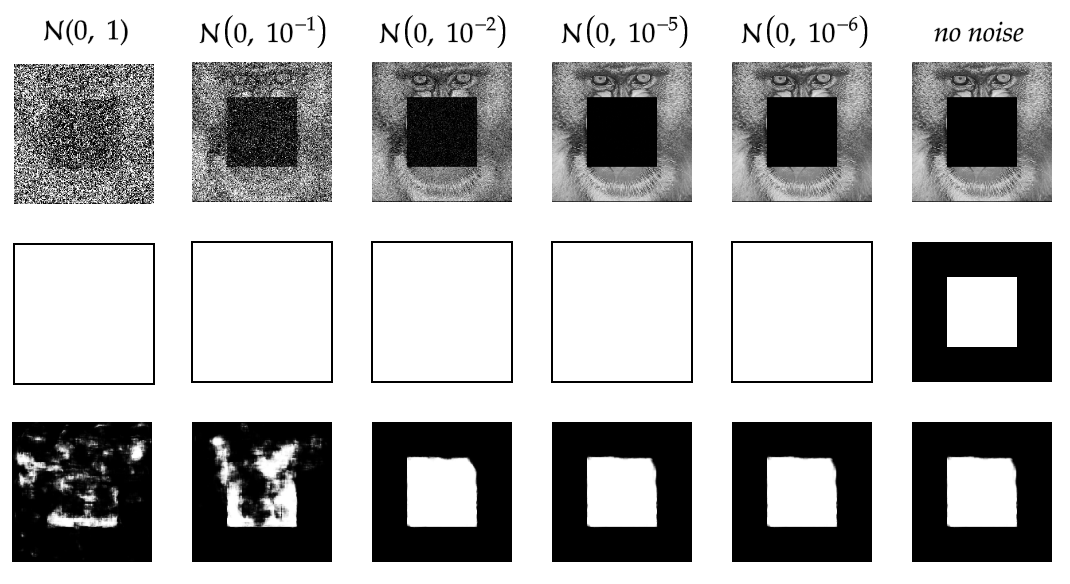}
  \caption{Robustness to additive noise for the \emph{baboon} image. Noise increases from right to left. From top to bottom: spliced image with added Gaussian noise; splice detection using block-level protection \cite{RePEc:gam:jmathe:v:7:y:2019:i:10:p:955-:d:275854}; and splice detection using our approach.}
  \label{fig:noise_robustness_results}
\end{figure}

The bottom of Figure~\ref{fig:noise_robustness_results} demonstrates the performance of our splice detector (Ours+N in the (\texttt{eu}) scheme) when increasing levels of i.i.d.~zero-mean Gaussian noise are added to the received image $\hat{I}$ before passing it to the decoder. In general we find that even without augmenting our training pipeline, our detection degrades gracefully until noise levels reach $10\%$ or more. As a point of reference, we also show the performance of a recent engineered method for block-level protection~\cite{RePEc:gam:jmathe:v:7:y:2019:i:10:p:955-:d:275854}, which impressively provides the additional capability of reconstructing the masked out-portion of a spliced image from its encoding, but fails to operate effectively even when the pixel values are affected by very small changes.



Additional experiments with other types of distortions are included in the supplement.


\subsection{Learned Residuals: Statistics and Structure}
\label{sec:statistics}

Minimizing the loss in Equation~(\ref{eqn:loss_function}) forces the encoder and decoder to learn a balance between positional accuracy and perceptual distortion. We can examine what they have learned by analyzing the statistics and structure of the residual images they create and use. 

\begin{figure}[h]
  \centering
\includegraphics[width=\columnwidth]{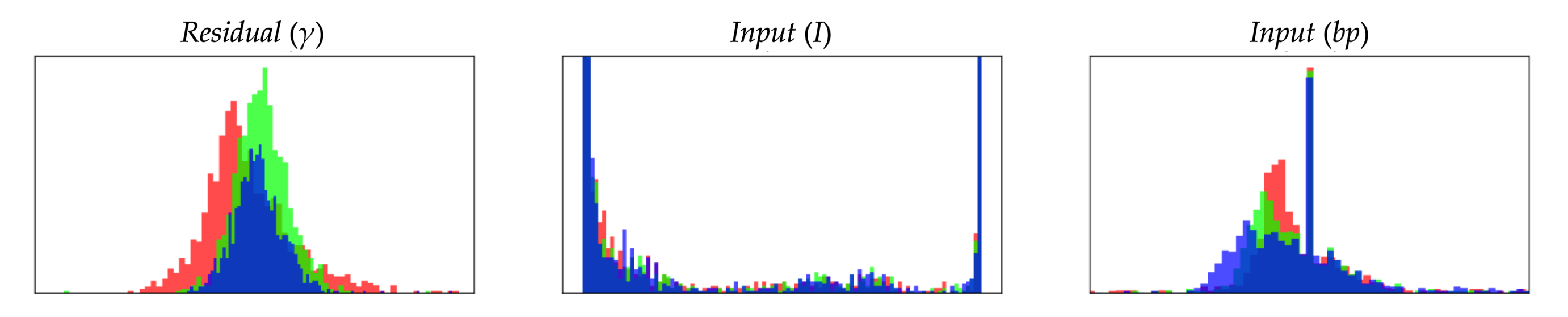}
    \caption{The RGB histogram of a residual image $\gamma$ (left) is different from that of its input image (middle) and is more similar to that of a bandpass-filtered input image (right).}
  \label{fig:residual-histogram}
\end{figure}

We start with first order statistics:  Figure~\ref{fig:residual-histogram}(a) shows normalized RGB histograms of the residual values $\gamma(x,y,c)$ created by the encoder for a single representative image that was not in the training set. (The residual histograms that are generated for all other input images are very similar.) The residual histograms are very different from the RGB histograms of the input image (b), and instead are more similar to those of a band-pass filtered version of the input image (c). The mean residual RGB value is $(0.002, 0.000, 0.004)$, which indicates that the overall color cast that it adds to the image is almost negligible. 

\begin{figure}[h]
  \centering
    \includegraphics[width=8cm]{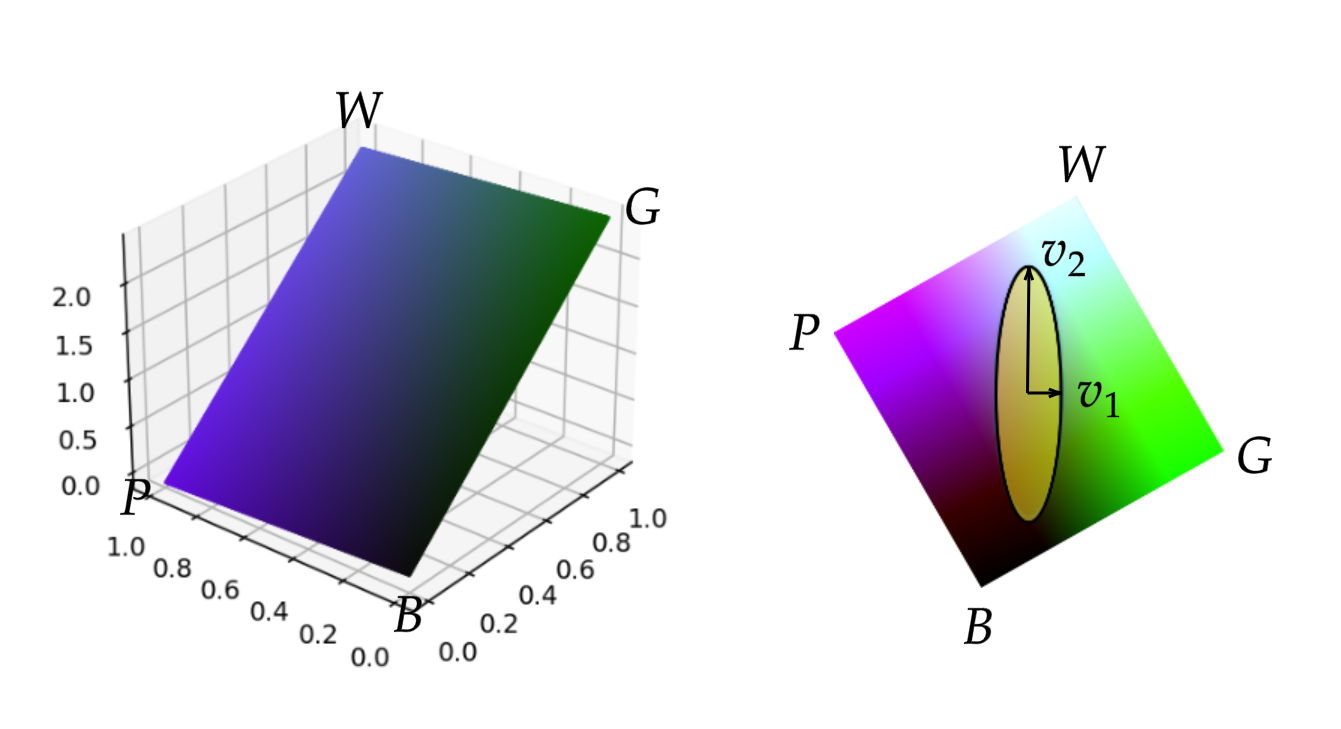}
    \caption{Residual colors $\gamma$ are highly concentrated near a plane in RGB color space. The plane is roughly spanned by the line of greys (in direction $v_2$) and by a line of chromaticities from green to purple (in direction $v_1$).}
  \label{fig:residual-plane}
\end{figure}

To explore the second-order statistics, we compute the mean residual color $\mu_\gamma$, subtract this mean from the per-pixel residual values of a single image, and compute the principal components of the resulting collection of three-vectors $(\gamma(x,y)-\mu_\gamma)\in\mathbb{R}^3$. Surprisingly, we find that $98\%$ of the variance in this set is explained by the first two components, indicating that the distribution of residual colors is highly concentrated near a plane in RGB space. This is in strong contrast to the distribution of colors in the input image, for which only $69\%$ of the variance is explained by the same plane. The ``residual plane'' is visualized in Figure~\ref{fig:residual-plane}. It comes close to containing the line of greys, and it contains a line of chromaticities spanning from roughly green to purple. This suggests that in the future, it may be worth exploring encoder architectures that create only bi-valued residuals and then combine them via 
\begin{equation}
\mu 
+a_\theta(x,y; 
\begin{bmatrix} I; \Psi \end{bmatrix})v_1
+b_\theta(x,y; 
\begin{bmatrix} I; \Psi \end{bmatrix})v_2,
\end{equation}
where $\mu,v_1,v_2\in\mathbb{R}^3$ are learned in conjunction with network parameters $\theta$. This would reduce encoder computation and could also help stabilize training.

\section{Conclusion and Societal Impact}

Our proof-of-concept model shows that it is possible,  using relatively simple CNNs, to steganographically inject localized positional signatures into images, and then to exploit these signatures for detecting crops and replacements in a way that can survive common distortions like tone adjustment and downsampling. This motivates exploring how to scale the approach to practical image sizes. It could help individuals and online platforms to know when an image can be certified as having been published by a trusted pool of content creators, and which parts of such an image, if any, have been substantially altered since publication.


We emphasize that our encoding can only help determine whether an image is faithful to the creator's intent and does not provide information about the nature of their intention. There is a risk that an authorised creator could abuse our system by intentionally encoding and publishing a fake or misleading image, thereby fooling individuals and platforms into thinking the image can be trusted. Similarly, an adversary could gain access to the encoder and use it to create an encoded image that is later falsely certified as trustworthy and having originated from the authorised pool of creators. Thus, any deployment would need to ensure that the encoder can only be accessed through a secure API, and that only certified, trustworthy agents can access it.

There is also a risk if an adversary gains uncontrolled access to the decoder. It is possible (but much less efficient) to create a stegapos embedding for a particular image without the encoder, by fixing the decoder and back-propagating the loss to the residual. An adversary could use this back-propagation attack as another way to create an image that is falsely trusted.

The risk of such backpropagation attacks could be reduced by requiring clients to query the decoder through a secure API---either to a server or to client-side confidential computing architecture---that prevents the use of differentiation for efficient backpropagation. Risk could be further reduced by limiting the temporal frequency of a client's queries to the decoder, or by preventing the client from making repeated queries using similar inputs, so that backpropagation with numerical differentiation becomes prohibitive in time.

Finally, we note that it may be possible for an adversary to effectively ``erase'' the stegapos-encoding from an image by creating and applying a large-but-imperceptible distortion that prevents the decoder from being able to detect any stegapos-encoding. This could preclude clients from being able to certify an image as trustworthy even though it may be a faithful rendition of an authorised creator's intent. Reducing this risk would require expanding the set of ``allowable distortions'' that the encoder and decoder are trained against, perhaps using an adversarial approach as proposed for conventional deep watermarking~\cite{luo2020distortion}.

\boldstart{Acknowledgements.} This work is supported by the National Science Foundation under Cooperative Agreement PHY-2019786 (an NSF AI Institute,  \href{http://iaifi.org}{http://iaifi.org}).

\bibliographystyle{plain}
\bibliography{main}

\newpage

\end{document}